\pdfoutput=1

\documentclass[11pt]{article}

\usepackage[final]{acl}

\usepackage{times}
\usepackage{latexsym}
\usepackage{amsmath}
\usepackage{amssymb}
\usepackage{xcolor}
\usepackage{algorithm,algorithmicx,algpseudocode}
\usepackage{wrapfig}
\usepackage{pgfplots}
\usepgfplotslibrary{groupplots}
\usepackage{multirow}
\usepackage{siunitx}
\usepackage{makecell}

\usepackage{enumitem}

\usetikzlibrary{plotmarks}
\usetikzlibrary{backgrounds}
\usetikzlibrary{fit}
\usetikzlibrary{decorations}
\usetikzlibrary{decorations.markings}
\usetikzlibrary{positioning}

\definecolor{bblue}{HTML}{4F81BD}
\definecolor{rred}{HTML}{C0504D}
\definecolor{ggreen}{HTML}{9BBB59}
\definecolor{ppurple}{HTML}{9F4C7C}
\definecolor{meta}{HTML}{4A7CAE}
\definecolor{meta1}{HTML}{A3A3A3}
\definecolor{meta2}{HTML}{43853E}

\definecolor{myblue}{RGB}{31,120,180}
\definecolor{mygreen}{RGB}{51,160,44}
\definecolor{myred}{RGB}{227,26,28}

\definecolor{iyellow}{RGB}{255,250,205}
\definecolor{ipurple}{RGB}{230,230,250}

\definecolor{upurple}{RGB}{155,89,182}
\definecolor{ublue}{RGB}{52,152,219}
\definecolor{ured}{RGB}{231,76,60}
\definecolor{udark}{RGB}{77,153,77}
\definecolor{ugreen}{RGB}{46,204,113}

\definecolor{bred}{RGB}{230,117,95}
\definecolor{bgreen}{RGB}{141,183,153}
\definecolor{bpurple}{RGB}{204,164,227}
\definecolor{bblue}{RGB}{117,193,196}

\newcommand{\camera}[1]{{\color{black}{#1}}}

\usepackage[T1]{fontenc}

\usepackage[utf8]{inputenc}

\usepackage{microtype}

\usepackage{inconsolata}

\usepackage{tikz}
\usetikzlibrary{shapes, arrows, positioning}

\usepackage{graphicx}
\usepackage{subcaption}
\usepackage{pifont}

\usepackage{booktabs}

\title{IIET: Efficient Numerical Transformer via 
Implicit Iterative Euler Method}

\author{
  \textbf{Xinyu Liu\textsuperscript{1*}},
  \textbf{Bei Li\textsuperscript{2*$\dagger$}},
  \textbf{Jiahao Liu\textsuperscript{2}},
  \textbf{Junhao Ruan\textsuperscript{1}},
  \\
  \textbf{Kechen Jiao\textsuperscript{3}},
  \textbf{Hongyin Tang\textsuperscript{2}},
  \textbf{Jingang Wang\textsuperscript{2}},
  \textbf{Xiao Tong\textsuperscript{1,4$\dagger$}}
  \textbf{Jingbo Zhu\textsuperscript{1,4}}
  \\
  \textsuperscript{1} \normalsize{NLP Lab, School of Computer Science and Engineering, Northeastern University, Shenyang, China} \\
  \textsuperscript{2} \normalsize{Meituan Inc.}
  \textsuperscript{3} \normalsize{Tsinghua University}
  \textsuperscript{4} \normalsize{NiuTrans Research, Shenyang, China} \\
  \normalsize{lxy1051493182@gmail.com, libei17@meituan.com} \\
  \normalsize{\{xiaotong, zhujingbo\}@mail.neu.edu.com}
}

\begin{document}
\maketitle

\renewcommand{\thefootnote}{\fnsymbol{footnote}}
\footnotetext[1]{These authors contributed equally to this work.}
\renewcommand{\thefootnote}{\dag}
\footnotetext[2]{Corresponding authors.}
\renewcommand{\thefootnote}{\arabic{footnote}}

\begin{abstract}

High-order numerical methods enhance Transformer performance in tasks like NLP and CV, but introduce a performance-efficiency trade-off due to increased computational overhead. Our analysis reveals that conventional efficiency techniques, such as distillation, can be detrimental to the performance of these models, exemplified by PCformer. To explore more optimizable ODE-based Transformer architectures, we propose the \textbf{I}terative \textbf{I}mplicit \textbf{E}uler \textbf{T}ransformer \textbf{(IIET)}, which simplifies high-order methods using an iterative implicit Euler approach. This simplification not only leads to superior performance but also facilitates model compression compared to PCformer. To enhance inference efficiency, we introduce \textbf{I}teration \textbf{I}nfluence-\textbf{A}ware \textbf{D}istillation \textbf{(IIAD)}. Through a flexible threshold, IIAD allows users to effectively balance the performance-efficiency trade-off. On lm-evaluation-harness, IIET boosts average accuracy by 2.65\% over vanilla Transformers and 0.8\% over PCformer. Its efficient variant, E-IIET, significantly cuts inference overhead by 55\% while retaining 99.4\% of the original task accuracy. Moreover, the most efficient IIET variant achieves an average performance gain exceeding 1.6\% over vanilla Transformer with comparable speed.

\end{abstract}

\section{Introduction}
The integration of advanced numerical ordinary differential equation (ODE) solvers into Transformer architectures \cite{vaswani2017attention} has spurred significant progress in natural language rocessing (NLP) \cite{li2022ode,li2024predictor,tong2025neural} and image synthesis \cite{ho2020denoising,lu2022dpm,lu2022dpm2,zheng2024dpm}. Leveraging high-order methods, particularly Predictor-Corrector (PC) schemes, within Transformer residual connections has demonstrated the capacity to enhance model learning without increasing parameter counts, offering a pathway to both performance and parameter efficiency \cite{li2022ode,li2024predictor}.

However, the promise of high-order PCformer \cite{li2024predictor} is often constrained by deployment inefficiencies. The inherent linear dependency in nested computations across layers during inference poses critical inference latency. A straightforward approach to mitigating this deployment bottleneck is Knowledge Distillation \cite{hinton2015distilling,kim2016sequence}. However, our preliminary experiments demonstrate that the inherent architectural discrepancy between the predictor and corrector within PCformer impedes effective knowledge transfer via distillation. Our empirical investigations reveal an obvious 54\% loss in performance advantage for distilled student models, even for those initialized with PCformer parameters.

Confronted with these deployment bottlenecks, we pivot towards architectural innovations grounded in numerical method principles. A naive yet seemingly logical initial approach might be to pursue uniformity in numerical methods between predictor and corrector, such as pairing explicit and backward Euler schemes. Similar attempts have been validated in previous studies~\cite{li2024predictor,zhao2024unipc}, where a high-order predictor combined with a single-step backward Euler method demonstrated promising results, particularly on smaller datasets. However, ensuring solution precision inherently requires iterative solvers to obtain the final solution, a process that shares the same merits as high-order methods. Building on this insight, we take a step further to explore whether an iterative corrector mechanism is equally critical for achieving both superior solution fidelity and unlocking genuine efficiency gains.  

\begin{figure*}[t]
  \centering
  \includegraphics[width=\textwidth]{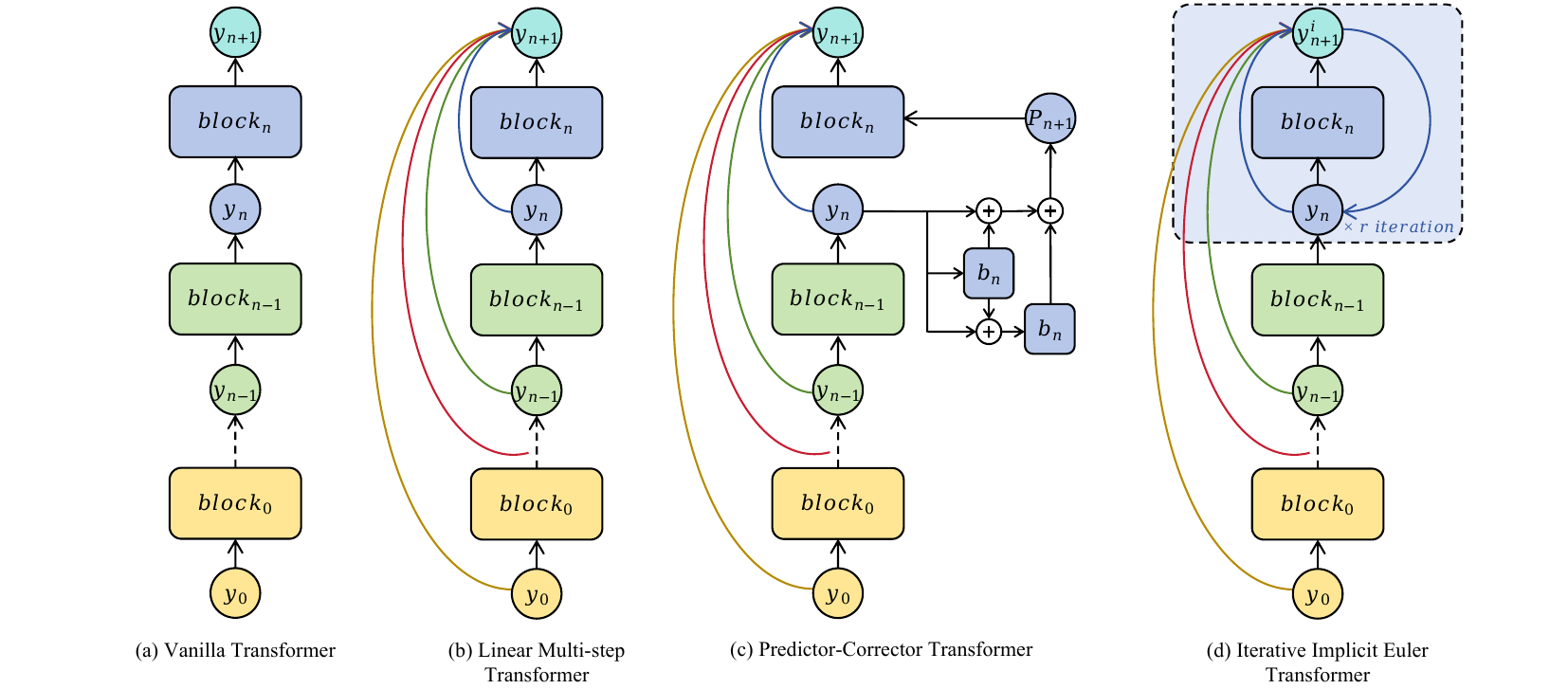}
  \caption{Architectural comparison: (a) Vanilla Transformer; (b) Linear multistep-enhanced Transformer; (c) PCformer with 2nd-order Runge-Kutta predictor and 1st-order Euler corrector; (d) Our proposed Iterative Implicit Euler Transformer (IIET). The iteration steps $r$ in IIET is configurable, with experimental validation determining $r=3$ as the optimal setting in this work. All blocks follow an identical computational procedure as the $block_n$.}
  \vspace{-3mm}
  \label{fig:main_arch}
\end{figure*}

To this end, we introduce the Iterative Implicit Euler Transformer (IIET). Concretely, in IIET, each iteration represents a computational step within an implicit Euler iterative solver, where multiple corrections to the initial prediction are made to ensure output precision. To further strengthen numerical stability, we also employ linear multi-step methods during each correction step. This architecture, detailed in Figure~\ref{fig:main_arch}d, is designed not only to achieve superior performance that scales with increasing iterations, exhibiting competitive results against PCformer, but also to be inherently compressible due to its iterative nature. \camera{Notably, our top-performing IIET models demonstrate a growing advantage over equivalent vanilla Transformers as they scale, achieving improvements of 2.4\% (340M), 2.9\% (740M), and a more substantial 5.6\% for the 1.3B model.}

In this way, we can effectively accelerate the inference of IIET via distillation techniques. Here, we further propose Iteration Influence-Aware Distillation (IIAD), a method inspired by structured pruning techniques \cite{men2024shortgpt,chen2024compressing}, to reduce dispensable iterations. Specifically, IIAD first assesses ``iteration influence'' by calculating input-output similarity for each iteration. The optimal number of iterations per layer is then determined according to a predefined influence threshold. Subsequently, a continued pre-training phase is employed to restore the model’s capabilities. This process enables users to tailor the iterative correction steps of the IIET model according to their computational budget, yielding efficient IIET variants. Experiments demonstrate that our efficient variant, E-IIET, reduces the inference computational cost of IIET by over 60\% while retaining \camera{98.7\%} of its performance. The lower bound of our efficient IIET variants not only outperforms the vanilla Transformer by an average of 1.6 points but also matches its inference efficiency, showcasing a significant advancement in both performance and deployment efficiency.

\section{Background}

We begin by establishing the connection between residual connections and the Euler method, and then discuss Transformer optimization strategies informed by advanced explicit and implicit numerical solutions of ODEs. Our work builds upon the standard Transformer architecture \cite{vaswani2017attention}, which comprises a stack of identical layers. For language modeling, each layer typically comprises a causal attention (CA) block and a feed-forward network (FFN) block. With residual connections, the output of each block can be formulated as $y_{n+1}=y_n + \mathcal{F}(y_n,\theta_n)$, where $\mathcal{F}(y_n,\theta_n)$ represents the transformation performed by either the CA or FFN block with parameters $\theta_n$.

\subsection{Euler Method in Residual Networks} 

The Euler method provides a linear approximation for first-order ODEs, defined as $y^{\prime}(t)=f(y(t),t)$ with an initial value $y(t_0)=y_0$. Given a step size $h$ where $t_{n+1}=t_n+h$, the method computes the subsequent value $y_{n+1}$ as:
\begin{equation}
    y_{n+1} = y_n + hf(y_n, t_n)
    \label{eq:euler_method}
\end{equation}
where $f(y_n, t_n)$ represents the rate of change of $y$, determined by its current value and time $t$. Notably, this formulation shares a structural similarity with residual networks, where a trainable function, $\mathcal{F}(\cdot)$, approximates these changes. Consequently, from an ODE perspective, residual connections can be interpreted as a first-order discretization of the Euler method. Although the success of residual connections highlights the benefits of the Euler method, its first-order nature introduces significant truncation errors~\cite{li2022ode,li2024predictor}, limiting the precision of $y_{n+1}$. Fortunately, more advanced numerical methods exist and have been successfully applied to neural networks.

\subsection{Advanced Numerical Transformers}

To improve the precision of $y_{n+1}$, the Runge-Kutta (RK) method offers a more accurate alternative. Inspired by the $o$-order RK method, the ODE Transformer~\cite{li2022ode} replaces residual connections with a RK process:
\begin{align}
    y_{n+1} &= y_{n} + \sum\nolimits_{i=1}^o \gamma_i \mathcal{F}_i \label{eq:odetransformer}\\
    \mathcal{F}_1 &= \mathcal{F}(y_n, \theta_n) \\
    \mathcal{F}_i &= \mathcal{F}(y_n + \sum\nolimits_{j=1}^{i-1} \beta_{ij} \mathcal{F}_j, \theta_n)
\end{align}
where $\mathcal{F}_i$ represents the $i^{th}$ order results computed by a shared transformer block $ \mathcal{F}(*, \theta_n)$. The coefficients $\gamma_i,\beta_{ij}$ are learnable parameters. This architecture effectively mitigates truncation error, leading to significant performance gains in generation tasks such as machine translation.

Compared to explicit numerical methods, implicit numerical methods typically offer higher precision and stability. The Predictor-Corrector (PC) method, using an explicit predictor for initial estimates and an implicit corrector for refinement, is a classic example. Recent work has demonstrated the benefits of integrating PC components into neural network architecture. PCformer~\cite{li2024predictor} employs an $o$-order RK predictor and a linear multi-step~\cite{wang2019learning} corrector, defined as:
\begin{equation} 
y_{p} = y_{n} + \sum\nolimits_{i=1}^o \gamma(1-\gamma)^{o-i}\mathcal{F}_i
\label{eq:predictor_network}
\end{equation}
\vspace{-5mm}
\begin{equation}
y_{n+1} = y_{n} + \alpha\mathcal{F}(y_p, \theta_n) + \sum_{i=n-2}^n\beta\tilde{\mathcal{F}}_i
\label{eq:corrector_network}
\end{equation}
where $\mathcal{F}_i$ shares the same meaning as in Eq.~\ref{eq:odetransformer} and $\tilde{\mathcal{F}}_i$ denotes the outputs of previous blocks. $\alpha,\beta,$ and $\gamma$ are learnable coefficients. Specifically, PCformer's predictor incorporates an Exponential Moving Average (EMA) to weight the contributions of different orders, while the corrector integrates previous block outputs for increased precision. PCformer achieves superior performance over the ODE Transformer and, to some extent, unifies structural paradigms for Transformers improved with implicit numerical methods. Our IIET can be interpreted as a specific instance within the PC paradigm, with a particular emphasis on the iterative corrector component.

\section{Iterative Implicit Euler Transformer}

In this section, we detail the theoretical foundation and core architectural design of the Iterative Implicit Euler Transformer (IIET).  Our approach leverages the inherent stability of the implicit Euler method, a cornerstone of numerical analysis, to address key challenges in deep sequence modeling.

\subsection{Iterative Implicit Euler Method}

The implicit Euler method, also known as the backward Euler method, is a foundational first-order implicit numerical technique celebrated for its robust stability properties, particularly advantageous in handling stiff systems \cite{finite2007siam}. Unlike its explicit counterparts, the implicit Euler method employs a backward difference quotient, formulated as:
\begin{equation}
    y_{n+1} = y_n + hf(y_{n+1}, t_{n+1}).
    \label{eq:implicit_euler_method}
\end{equation}
The implicit nature of Eq.~\ref{eq:implicit_euler_method}, where the computation of $y_{n+1}$ depends on its value at the same time step $t_{n+1}$, inherently requires iterative solvers from numerical analysis to obtain a solution. Specifically, in traditional numerical methods for solving such implicit equations, Newton's iteration is frequently employed due to its quadratic convergence rate and robustness \cite{zhang2017polynet,shen2020implicit,kim2024bert}. However, within the context of neural sequence modeling, where computational efficiency and architectural simplicity are often prioritized, we propose to investigate the efficacy of a simpler alternative: fixed-point iteration \cite{rhoades1976comments}. While prior works like \citet{li2024predictor} have utilized explicit methods for initial approximations followed by a single backward Euler correction, the potential of iterative refinement within the implicit corrector remains largely unexplored.

Thus, challenging the implicit assumption that a strong predictor is sufficient for high precision \citep{li2024predictor}, we propose the central hypothesis that iterative refinement inside the implicit corrector constitutes a pivotal mechanism for enhancing solution fidelity.  We argue that a single-step correction inherently limits the achievable accuracy, particularly when modeling intricate sequence dynamics and seeking high-fidelity representations of $y_{n+1}$. Consequently, this work rigorously investigates whether leveraging iterative solutions within the implicit corrector can translate to demonstrable gains in downstream model performance.

Intriguingly, our empirical findings reveal that computationally efficient fixed-point iteration yields surprisingly high precision, 
particularly within our neural sequence modeling framework. Our proposed Iterative Implicit Euler (IIE) method commences with an initial approximation, $y_{n+1}^0$, derived from an explicit Euler step. This initial estimate is then iteratively refined through $r$ fixed-point iterations as defined below:
\begin{align}
    y_{n+1}^0 &= y_{n} + hf(y_n, t_n)
    \label{eq:implicit_euler_method_initial} \\
    y_{n+1}^i &= y_{n} + hf(y_{n+1}^{i-1}, t_{n+1}), \quad i \in [1.. r].
    \label{eq:implicit_euler_method_iteration}
\end{align}
The final approximation $y_{n+1}$ is thus given by $y_{n+1}^r$, representing the output of the $r^{th}$ iteration.

The IIE method, while formally retaining its first-order numerical accuracy, achieves a significant enhancement in the approximation of $y_{n+1}$ through iterative refinement. This iterative process engenders a structured form of nested computations that superficially resemble higher-order methods, albeit through a fundamentally distinct mechanism rooted in repeated fixed-point iterations. Acknowledging the increased computational cost, the inherent structural regularity of IIE, predicated solely on the preceding iteration's output, emerges as a crucial enabler for inference efficiency optimizations, as detailed in Section~\ref{sec:IIAD}. This carefully engineered balance between iteratively enhanced precision and structural simplicity underpins the design philosophy of the IIET architecture.

\subsection{Model Architecture}

\begin{algorithm}[htbp]
\small
\caption{Iterative Implicit Euler Paradigm}
\label{alg:iiet_detailed}
\begin{algorithmic}[1]
  \Procedure{IIET Block}{$\mathbf{y_n}$, $\theta_n$, $\mathbf{L}$, $r$}

    \State $\mathbf{L}$ is the global stack for historical hidden states
    \State $f_n^0 \leftarrow \mathcal{F}(\mathbf{y_n}, \theta_n)$ \Comment{Compute initial value}
    \State $\mathbf{L}$.append($f_n^0$)  \Comment{Store the initial context}

    \For{$i \leftarrow 0$ \textbf{to} $r-1$} \Comment{Refinement loop}

        
        \State Compute $\mathbf{y_{n+1}^{i+1}}$ using $\mathbf{L}$ via the update rule
        
        \State $f_n^{i+1} \leftarrow \mathcal{F}(\mathbf{y_{n+1}^{i+1}}, \theta_n)$ \Comment{Compute refined value}
        \State $\mathbf{L}$.update($f_n^i \rightarrow f_n^{i+1}$) \Comment{Update the context $\mathbf{L}$}
    \EndFor

     \State Compute $\mathbf{y_{n+1}^r}$ using the final context $\mathbf{L}$
    \State \textbf{return} $\mathbf{y_{n+1}^r}$ \Comment{Return the final layer output}

  \EndProcedure
\end{algorithmic}
\end{algorithm}

Building on the IIE method, we propose the Iterative Implicit Euler Transformer (IIET) as a foundational architecture for sequence modeling, particularly for large language models. Adopting the LLaMA architecture~\cite{touvron2023llama2} (Transformer++), IIET consists of $N$ stacked transformer decoder layers. Each layer comprises a causal attention module followed by a feedforward module, and employs rotary positional encoding~\cite{su2024roformer}, SiLU activation~\cite{shazeer2020glu}, and RMS normalization~\cite{zhang2019root}. Given an input sequence $x=x_1,...,x_L$ of length $L$, the initial input embeddings are represented as $X^0=[x_1,...,x_L]\in \mathbb{R}^{L \times d_{\text{model}}}$, where $d_{\text{model}}$ is the hidden dimension. The output of each subsequent layer is then computed as $X^n=\text{Decoder}(X^{n-1})$, for $n\in [1,N]$.

The key distinction between IIET and Transformer++ lies in IIET's integration of the IIE method within each decoder layer (Figure~\ref{fig:main_arch}). Unlike Transformer++, which directly computes the layer's output using a single Euler step (standard residual), IIET employs an iterative refinement process. Specifically, IIET first estimates an initial value, $y_{n+1}^0$, via a single Euler step (Eq.~\ref{eq:implicit_euler_method_initial}):
\begin{equation}
    y_{n+1}^0 = y_n + \mathcal{F}(y_n,\theta_n)
    \label{eq:icformer_output_euler}
\end{equation}
\noindent where $\mathcal{F}(*,\theta_n)$ represents the $n^{th}$ transformer layer with parameters $\theta_n$. This initial estimate in IIET corresponds to the direct output of each layer in Transformer++.

In the subsequent iterations, our preliminary experiments suggest that incorporating outputs from previous layers, similar to Transformer-DLCL~\cite{wang2019learning}, can enhance the performance. We thus modify Eq.~\ref{eq:implicit_euler_method_iteration} as follows:
\begin{equation}
    y_{n+1}^i = y_n + \alpha_n \mathcal{F}(y_{n+1}^{i-1},\theta_n) + \sum_{j=0}^{n-1} \alpha_j \tilde{\mathcal{F}}_j
    \label{eq:icformer_iteration}
\end{equation}
\noindent where $i \in [1.. r]$ denotes the iteration step, $\tilde{\mathcal{F}}_j$ represents the output of the previous layers $j$, and $\alpha$ represents learnable layer merge coefficients. 

\camera{Algorithm~\ref{alg:iiet_detailed} further details the computational flow within a single IIET layer. Specifically, the matrix $\mathbf{L}$ stores the hidden states from previously computed layers, thereby providing the necessary historical context. Within a single block, an initial estimate of the output, denoted as $y_{n+1}^0$, is generated and then iteratively refined. Each iteration $i$ updates this estimate to $y_{n+1}^i$ by leveraging the function $\mathcal{F}$ and the context $\mathbf{L}$. This fixed-point iteration process ensures that the hidden state progressively converges to a more precise and stable final output, $y_{n+1}^r$.}

\definecolor{ublue}{RGB}{39,89,167}
\definecolor{ured}{RGB}{172,21,28}
\definecolor{ugreen}{RGB}{34,139,34}
\definecolor{upurple}{RGB}{94,53,177}
\begin{figure}[!t]
\centering
\begin{tikzpicture}
    \scriptsize
    \begin{groupplot}[
        group style={
            group size=2 by 1,
            horizontal sep=10pt,
            y descriptions at=edge left,
            x descriptions at=edge bottom,
        },
        width=0.30\textwidth,
        height=0.25\textwidth,
        xtick={1,2,...,4},
        ytick=\empty,
        grid style=dashed,
        ylabel={Perplexity},
        ylabel style={font=\scriptsize,yshift=-3.8em,align=center},
        y tick style={opacity=0},
        y tick label style={font=\tiny},
        ymajorgrids=true,
        xmajorgrids=true,
        tick align=inside,
        legend pos=outer north east,
        yticklabel style={/pgf/number format/precision=2,/pgf/number format/fixed zerofill},
        legend style={yshift=-3.5em,xshift=-11.1em,legend cell align=left,legend plot pos=right,fill opacity=0.5},
        xmin=1,xmax=4,
        enlargelimits=0.15,
        nodes near coords,
        point meta=explicit symbolic,
    ]

    \nextgroupplot[
        ymin=66.5,ymax=69,
        width=0.31\textwidth,
        title={IIET 55M},
        title style={font=\scriptsize,yshift=-0.6em},
        xlabel={}, 
    ]
    \addplot [sharp plot,ublue,mark=otimes*,mark size=1.5pt,thick,line width=0.6pt,mark options={fill=white,draw=ublue,line width=0.5pt}]
        table[row sep=crcr, meta=label] {
        x    y    label\\
        1    68.42  68.4\\
        2    67.38  67.4\\
        3    66.63  66.6\\
        4    67.35  67.4\\
        };
    \addplot [dashed, upurple, thick] coordinates {(0,69) (5,69)};
    \node[anchor=south east, text=upurple, font=\scriptsize] at (axis cs:4,69) {77.6};
    \addplot [dashed, ugreen, thick] coordinates {(0,68.2) (5,68.2)};
    \node[anchor=south east, text=ugreen, font=\scriptsize] at (axis cs:4,68.2) {68.2};
    
    \nextgroupplot[
        ymin=25,ymax=26.5,
        width=0.31\textwidth,
        title={IIET 340M},
        title style={font=\scriptsize,yshift=-0.6em},
        xlabel={}, 
    ]
    \addplot [sharp plot,ured,mark=otimes*,mark size=1.5pt,thick,line width=0.6pt,mark options={fill=white,draw=ured,line width=0.5pt}]
        table[row sep=crcr, meta=label] {
        x    y    label\\
        1    25.96  26.0\\
        2    25.49  25.5\\
        3    25.02  25.0\\
        4    25.19  25.2\\
        };
    \addplot [dashed, upurple, thick] coordinates {(0,26.5) (5,26.5)};
    \node[anchor=south east, text=upurple, font=\scriptsize] at (axis cs:4,26.5) {28.2};
    \addplot [dashed, ugreen, thick] coordinates {(0,25.7) (5,25.7)};
    \node[anchor=south east, text=ugreen, font=\scriptsize] at (axis cs:4,25.7) {25.7};

    \end{groupplot}
    
    \node[below,font=\scriptsize,yshift=-0.1em] at (current bounding box.south) {iteration steps $r$};
    
\end{tikzpicture}
\caption{PPL on the Wikitext test set for 55M and 340M IIET across varying iteration steps $r$. Dashed lines indicate \textcolor{upurple}{Transformer++} and \textcolor{ugreen}{PCformer} performance at corresponding parameter scales. Note that IIET's FLOPs is nearly $r+1$ times of Transformer++.}
\vspace{-3mm}
\label{fig:judge_iter_count}
\end{figure}
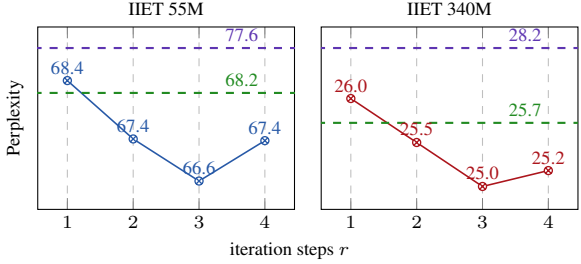

\subsection{Experimental Setups}
\label{sec:exp_results}
\begin{table*}[t!]
\centering
\small
\addtolength{\tabcolsep}{-2.5pt}    
\begin{tabular}{l l|cc|cccccc|c}
\toprule
&   & \textbf{Wiki.}  &  \textbf{LMB.} &  \textbf{LMB.} & \textbf{PiQA} &    \textbf{Hella.} & \textbf{SCIQ} &  \textbf{ARC-c} & \textbf{Wino.} &  \textbf{Avg.}  \\
\textbf{Scale} & \textbf{Model}  & ppl $\downarrow$  &  ppl $\downarrow$  &  acc $\uparrow$  & acc\_norm $\uparrow$ &   acc\_norm $\uparrow$  & acc $\uparrow$  & acc\_norm $\uparrow$ & acc $\uparrow$ &  $\uparrow$ \\



\midrule
\vspace{-1.2em} \\
\multicolumn{11}{l}{\textit{Pre-training Phase}} \\
\vspace{-1.2em} \\
\midrule

\textit{340M Params} & Transformer++ & 28.2 & 78.3 & 28.9 & 64.3 & 34.2 & 76.0 & 23.6 & 51.9 & 46.5  \\
\textit{16B Tokens} & PCformer & 25.7 & 47.0 & 33.1 & 64.9 & 36.3 & 77.5 & \textbf{24.7} & \textbf{53.3} & 48.3 \\
& IIET & \textbf{25.0} & \textbf{30.5} & \textbf{37.1} & \textbf{65.2} & \textbf{36.9} & \textbf{79.4} & 23.9 & 51.0 & \textbf{48.9}\\

\midrule 

\textit{740M Params} & Transformer++ & 23.3 & 34.8 & 36.1 & 66.4 & 38.4 & 78.6 & \textbf{24.5} & 50.2 & 49.0  \\
\textit{30B Tokens} & PCformer & 21.2 & 22.0 & 41.0 & 66.3 & 41.3 & 82.0 & 23.3 & 51.2 & 50.9 \\
& IIET & \textbf{20.7} & \textbf{21.1} & \textbf{41.2} & \textbf{68.9} & \textbf{42.5} & \textbf{82.1} & 23.8 & \textbf{53.1} & \textbf{51.9}\\

\midrule 

\textit{1.3B Params} & Transformer++ & 16.3 & 11.8 & 51.6 & 71.0 & 51.7 & 86.7 & 28.1 & 54.6 & 57.2  \\
\textit{100B Tokens} & PCformer & \textbf{14.0} & 7.9 & 59.6 & \textbf{73.8} & 60.0 & \textbf{90.7} & 31.7 & \textbf{61.7} & \textbf{62.9} \\
& IIET & \textbf{14.0} & \textbf{7.8} & \textbf{59.8} & 73.7 & \textbf{60.5} & 88.6 & \textbf{32.3} & 61.6 & 62.8 \\

\midrule
\vspace{-1.2em} \\
\multicolumn{11}{l}{\textit{Iteration Influence-Aware Distillation Phase}} \\
\vspace{-1.2em} \\
\midrule

\textit{340M Params} & Distil PCformer & 27.2 & 50.4 & 32.2 & \textbf{64.6} & 34.9 & 78.0 & 24.7 & 51.3 & 47.6 \\
\textit{5B Tokens} & Lower Bound & 27.0 & 34.6 & 36.1 & 64.0 & 35.0 & \textbf{80.7} & 23.0 & 51.5 & 48.4 \\
& E-IIET & \textbf{25.7} & \textbf{30.9} & \textbf{37.4} & 64.4 & \textbf{35.8} & 80.4 & \textbf{23.5} & \textbf{52.1} & \textbf{48.9} \\


\midrule 

\textit{740M Params} & Distil PCformer & 22.5 & 29.5 & 37.4 & 66.8 & 39.2 & 80.0 & 23.2 & 50.9 & 49.6 \\
\textit{10B Tokens} & Lower Bound & 23.0 & 29.9 & 37.6 & 67.4 & 38.7 & 79.7 & \textbf{25.2} & \textbf{53.0} & 50.3 \\
& E-IIET & \textbf{21.2} & \textbf{24.2} & \textbf{40.1} & \textbf{68.5} & \textbf{41.0} & \textbf{81.0} & 24.6 & 52.4 & \textbf{51.3} \\

\midrule 

\textit{1.3B Params} & Lower Bound & 15.9 & 9.3 & 57.0 & 72.7 & 56.7 & 87.7 & 29.6 & 59.1 & 60.5 \\
\textit{30B Tokens} & E-IIET & \textbf{14.8} & \textbf{8.9} & \textbf{57.5} & \textbf{73.2} & \textbf{58.4} & \textbf{88.6} & \textbf{30.7} & \textbf{59.4} & \textbf{61.3} \\



%
\bottomrule
\end{tabular}
\addtolength{\tabcolsep}{2.5pt}    
\centering
\caption{Comparison of results between our models and baselines in the \textit{Pre-training Phase} and \textit{Iteration Influence-Aware Distillation Phase}. The individual task performance is via zero-shot. We report the main results on the same set of tasks reported by \citet{Gu2023Mamba}. The last column shows the average over all benchmarks that use (normalized) accuracy as the metric. \textbf{Bold} values represent the best results in each set.}
\vspace{-3mm}
\label{tab:main_results}
\end{table*}

\paragraph{Baselines.} We evaluate IIET's performance against two strong baselines: Transformer++~\cite{touvron2023llama} and PCformer~\cite{li2024predictor}. Transformer++ adopts the LLaMA architecture. PCformer employs a 2nd-order Runge-Kutta predictor and a linear multi-step corrector~\footnote{We also explored a 4th-order Runge-Kutta predictor and more complex correctors, but these increased training costs without substantially improving performance.}. \camera{We train all models from scratch at three parameter scales: 340M, 740M, and 1.3B. All models are trained on the same dataset with identical token counts to ensure controlled comparison.} Detailed training hyperparameter settings can be found in Appendix~\ref{sec:pretrain_detail}.

\paragraph{Datasets and Evaluation Metrics.}
Our models are pre-trained on SlimPajama~\cite{soboleva2023slimpajama} and tokenized using the LLaMA2 tokenizer~\cite{touvron2023llama}. From the original 627B-token dataset, \camera{we sample 16B, 30B and 100B tokens for training the 340M, 740M and 1.3B parameter models, respectively.} For comprehensive evaluation, we assess perplexity (PPL) on Wikitext (Wiki.)~\cite{merity2016pointer} and consider several downstream tasks covering common-sense reasoning and question answering: LAMBADA (LMB.)~\cite{paperno2016lambada}, PiQA~\cite{bisk2020piqa}, HellaSwag (Hella.)~\cite{zellers2019hellaswag}, WinoGrande (Wino.)~\cite{sakaguchi2021winogrande}, ARC-Challenge (ARC-c)~\cite{clark2018think}, and SCIQ~\cite{welbl2017crowdsourcing}. We report PPL on Wikitext and LAMBADA; length-normalized accuracy on HellaSwag, ARC-Challenge, and PiQA; and standard accuracy on the remaining tasks. All evaluations are conducted using the lm-evaluation-harness~\cite{gao2021framework}. \camera{In addition to language modeling, we also conduct experiments on machine translation and summarization tasks, with detailed results in Appendix~\ref{sec:results_on_other_tasks}.}

\label{sec:ablation_r}
\begin{figure}[t]
  \centering
  \includegraphics[width=0.49\textwidth]{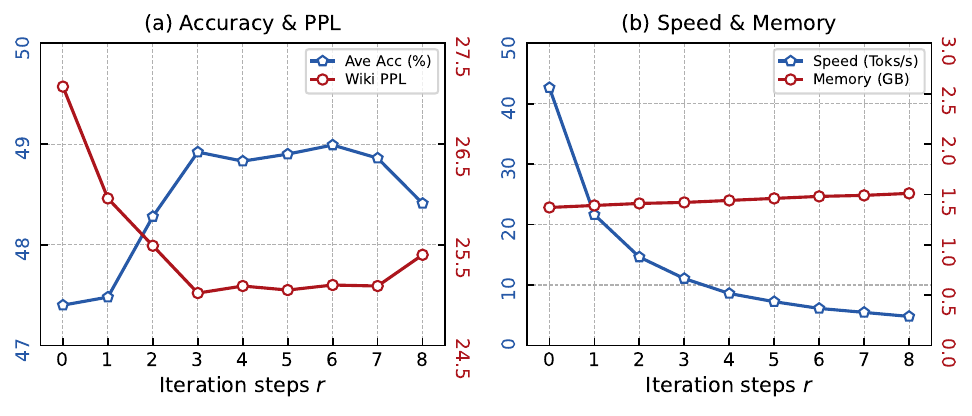}
  \caption{Ablation study on iteration steps $r$: (a) Impact on model performance. (b) Corresponding effects on inference speed and VRAM utilization.}
  \vspace{-3mm}
  \label{fig:ablation}
\end{figure}

\subsection{Experimental Results}

\paragraph{Iteration Steps.}
To identify the optimal iteration steps $r$, we first apply varying $r$ values to the 340M IIET model and a smaller 55M parameter variant (detailed in Appendix~\ref{sec:pretrain_detail}). All models were evaluated on Wikitext test set. As illustrated in Figure~\ref{fig:judge_iter_count}, which showcases the benefit of iterative correction, IIET's performance exceeds PCformer at $r=2$ and achieves its peak at $r=3$. Therefore, we adopt $r=3$ in this work.

\paragraph{Results.}
\camera{IIET's advantages are clearly demonstrated on LLM evaluation benchmarks. As shown in Table~\ref{tab:main_results} (\textit{Pre-training Phase}), IIET consistently surpasses Transformer++ at comparable parameter scales. At the 340M scale, IIET achieves a mean accuracy 2.4 points higher than Transformer++ across six challenging subtasks. This performance gap widens with model size, reaching 2.9 points at 740M and a substantial 5.6 points at 1.3B parameters. Moreover, IIET's performance also matches or exceeds that of PCformer across all tested scales, demonstrating the advantages of the iterative correction paradigm. This strong scaling behavior, consistent with the findings in \citet{li2024predictor}'s work, confirms the robust scalability of IIET and showcases its potential with larger models and datasets.}

\subsection{Analysis}

\label{sec:iiet_ana}

\paragraph{Ablation Study on Iteration Steps.}
A key question concerning IIET is whether its performance improves monotonically with an increasing number of iterative correction steps. To investigate this, we conducted an ablation study on the 340M IIET model, varying the number of iteration $r$.~\footnote{In the case where $r=0$, IIET is structurally the same as the DLCL Transformer.} As illustrated in Figure~\ref{fig:ablation}a, performance initially improves with increasing $r$. However, beyond a certain threshold, further increases in $r$ lead to a plateau in performance gains. This suggests that the iterative refinement process guides the final representation towards a more precise ODE solution, but with diminishing returns after optimal convergence. Detailed downstream results can be found in Appendix~\ref{sec:analysis_iter_step_detail}. FurtherMore, to assess the impact of $r$ on inference efficiency, we measured the autoregressive generation throughput of IIET variants on a single A100 GPU. Figure~\ref{fig:ablation}b shows that while IIET's inference speed substantially declines with increasing $r$, its VRAM footprint remains largely unaffected as it incurs no extra parameters.

\begin{figure*}[htbp]
  \centering
  \includegraphics[width=\textwidth]{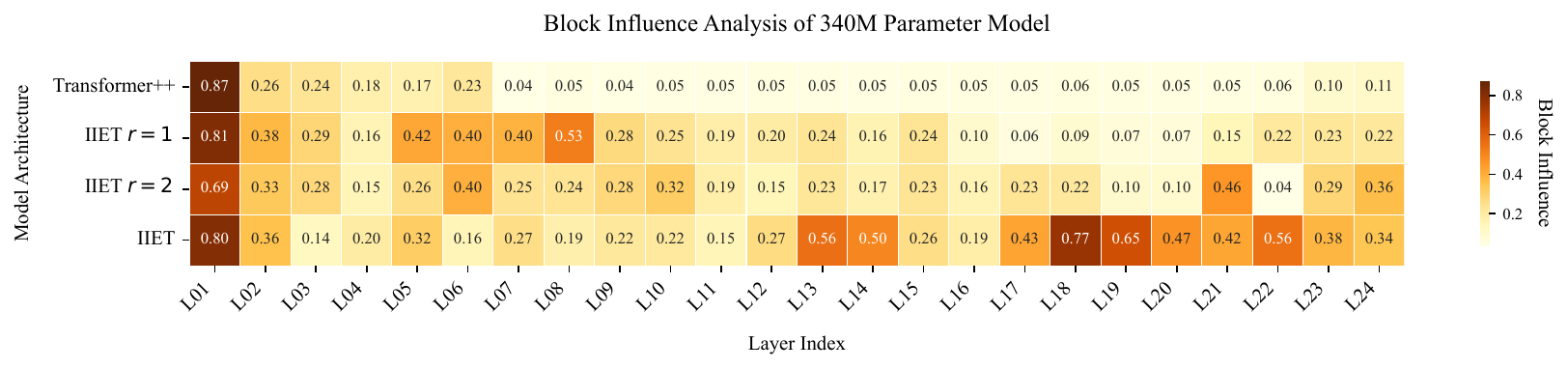}
  \caption{Distribution of Block Influence (BI) for Transformer++ and IIET models with varying iteration steps $r$. Higher BI values indicate lower model redundancy.}
  \vspace{-3mm}
  \label{fig:analysis_block_influence}
\end{figure*}

\begin{table}[t]
    \centering
    \small
    \setlength{\tabcolsep}{1.6pt}
    \begin{tabular}{l ccccccc}
    \toprule
    \textbf{Model} & {\textbf{LMB.}} & {\textbf{PiQA}} & {\textbf{Hella.}} & {\textbf{SCIQ}} & {\textbf{ARC-c}} & {\textbf{Wino.}} & {\textbf{Avg.}} \\
    \midrule
    IIET & 37.1 & 65.2 & 36.9 & 79.4 & 23.9 & 51.0 & 48.9 \\
    Trans WS & 30.7 & 63.1 & 34.4 & 75.7 & 23.2 & 50.4 & 46.3 \\
    Trans 1.3B & 37.3 & 65.7 & 37.6 & 78.6 & 23.7 & 51.5 & 49.0 \\
    \bottomrule
    \end{tabular}
    \caption{Performance comparison of models with FLOPs comparable to the 340M IIET.}
    \vspace{-3mm}
    \label{tab:analysis_comparison_flops}
\end{table}

\paragraph{Comparison with Equal FLOPs.} Given that IIET's iterative correction adds FLOPs (to approximately four times that of Transformer++ when $r=3$), we aimed for a performance comparison under equivalent computational budgets. Thus, we trained a 1.3B Transformer++ model on identical 16B training data. The results in Table~\ref{tab:analysis_comparison_flops} show that IIET performs comparably to the much larger Transformer++ but with substantially fewer parameters, thereby reducing memory and training overhead. \camera{To ensure a fair comparison based on model size, we compressed the 1.3B Transformer++ model using pruning and quantization to align with IIET's storage requirements. The results of this evaluation are detailed in Appendix~\ref{sec:comparison_with_compressed_models}.} Moreover, models with exactly matched parameter scale and FLOPs were benchmarked. Since IIET's architecture closely resembles weight-sharing methods, we established a naive weight-sharing baseline: the Transformer++ model's depth was quadrupled, with weights shared every four layers, namely Trans WS. As shown in Table~\ref{tab:analysis_comparison_flops}, this simple weight-sharing approach alone does not yield performance gains, highlighting the crucial contribution of IIET's implicit iterative solver-based design to its enhanced performance.

\paragraph{Parameter Redundancy of IIET.} 
We hypothesize that the iterative correction process of IIET enhances learning efficiency and reduces parameter redundancy. To investigate this, we used Block Influence (BI)~\cite{men2024shortgpt} to measure layer redundancy in IIET and Transformer++. BI assesses the influence of each model block on the hidden state by measuring the similarity between its input and output; lower similarity indicates a higher influence. Specifically, the BI of a Transformer block is calculated as:
\begin{equation}
    \text{BI}_i = 1 - \mathbb{E}_{\mathbf{H}, t} \frac{\mathbf{H}_{i, t}^T \mathbf{H}_{i+1, t}}{||\mathbf{H}_{i, t}||_2 ||\mathbf{H}_{i+1, t}||_2}
    \label{block_influence}
\end{equation}
where $\mathbf{H}_{i,t}$ represents the $t^{th}$ row of the $i
^{th}$ layer's input hidden states. We randomly sampled 5,000 text segments from Wikitext to calculate the BI of each model. As shown in Figure~\ref{fig:analysis_block_influence}, the influence of IIET's blocks increases significantly with iteration steps, demonstrating higher layer utilization. This also indicates that the learning potential of existing large-scale language models remains under-exploited.

\section{Iteration Influence-Aware Distillation}

\label{sec:IIAD}

\begin{figure*}[t]
  \centering
  \includegraphics[width=\textwidth]{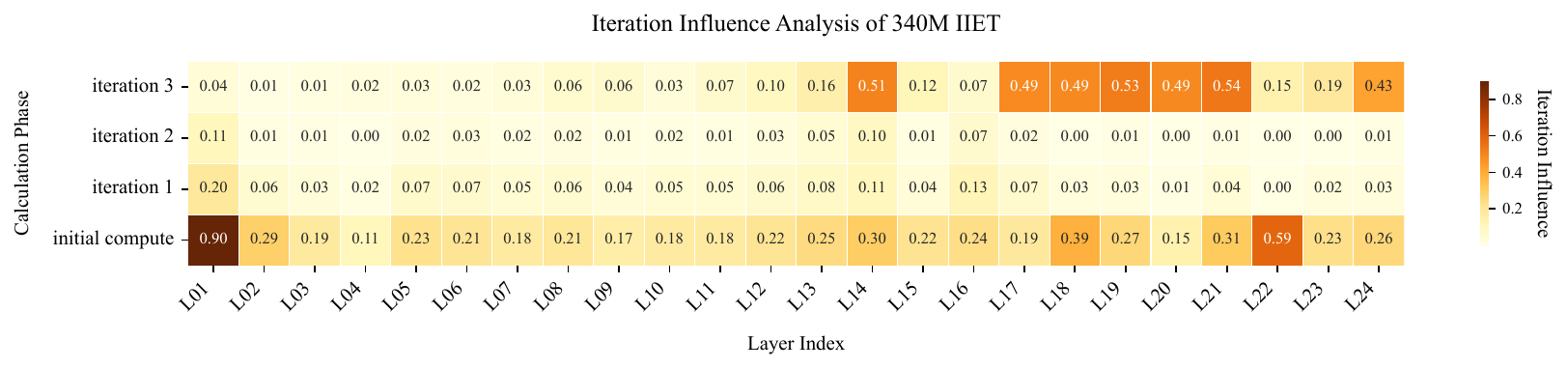}
  \caption{\textbf{Iteration Influence} within each layer of the 340M IIET model. Deeper colors indicate larger hidden state changes after this iteration. The 740M IIET results are presented in Appendix~\ref{sec:740m_iiet_heatmap} due to space constraints.}
  \vspace{-3mm}
  \label{fig:fine_grained_bi_heatmap}
\end{figure*}

While IIET achieves strong downstream task performance, its iterative structure introduces computational overhead that curtails inference speed. This added latency is particularly non-negligible for autoregressive generation in large language models. To enhance IIET's inference efficiency without performance loss, we explore whether continuous pre-training combined with distillation can enable fewer forward passes, ideally a single one, to yield outputs equivalent to those from the complete, multi-step iterative correction process. To this end, we analyze the impact of each iterative correction step on the hidden state within each block. Surprisingly, Figure~\ref{fig:fine_grained_bi_heatmap} shows that not all layers require the same number of iteration steps to achieve accurate output, with deeper layers benefiting more from additional iterative corrections, which is potentially due to the varying roles layers play in the Transformer's representation-building process.

\subsection{Methodology}

\label{sec:iiad_method}

In this section, we propose Iteration Influence-Aware Distillation (IIAD). IIAD first analyzes the iterative process of a pre-trained IIET, identifying and eliminating non-essential iterative computations to yield an efficient variant, E-IIET. Subsequently, a layer-wise self-distillation phase restores the performance of E-IIET.

\paragraph{Iteration Influence.}

Iteration influence employs a computational methodology similar to block influence; however, its calculation is performed specifically within individual IIET blocks. For a given $n^{th}$ block, we consider its input $y_n$ and the output $y_{n+1}^i$ of each internal iteration $i$. The pairwise differences between these representations are calculated using Eq.~\ref{block_influence} to obtain the iteration influence values. Based on these values and a specified computational budget, users can determine the number of iteration steps to retain per block.

In this work, we primarily investigate two designs for efficient IIET variants: \ding{182} \textbf{Lower Bound}: Each layer performs only a single forward pass, establishing a performance lower bound for efficient IIET. \ding{183} \textbf{E-IIET}:  This variant establishes a threshold using the minimum of the initial iteration influence values computed in each layer. Consequently, iteration steps with influence scores below this threshold are omitted, preserving each layer's initial computation and essential iteration steps. Specifically, E-IIET reduces the number of iteration steps from a baseline of 72 to 15 in the 340M variant and 23 in the 740M variant. 

\paragraph{Iteration Influence-Aware Distillation.}
In the continuous pre-training stage, we employ a warm-start initialization strategy, directly inheriting parameters from the pre-trained IIET model to retain knowledge acquired during its initial pre-training phase. To enable efficient IIET variants (e.g., E-IIET) to approximate the precise output representations of the full IIET, we utilize a fine-grained, block-specific knowledge distillation framework incorporating two complementary losses: \textbf{1) Mean Squared Error (MSE) Loss}: For each block, an MSE loss encourages E-IIET to mimic the refined hidden states produced by the full IIET. This loss is computed as:
\begin{equation}
    \mathcal{L}_\text{MSE} = \frac{1}{n}\sum_{i=1}^n \| \mathbf{h}_i^\text{IIET} - \mathbf{h}_i^{\text{E-IIET}} \|_2^2
\end{equation}
\noindent where $\mathbf{h}_i$ are the hidden state outputs of the $i^{th}$ block. \textbf{2) Kullback-Leibler (KL) Loss}: To further align prediction behavior, we compute the KL divergence between the final output probability distributions of the full IIET and E-IIET:
\begin{equation}
    \mathcal{L}_\text{KL} = D_\text{KL}\left( p(\mathbf{z}^\text{IIET} / \tau) \parallel p(\mathbf{z}^{\text{E-IIET}} / \tau) \right)
\end{equation}
\noindent where $\mathbf{z}$ represent the output logits and $\tau$ is the distillation temperature. By combining these two distillation losses with Cross-Entropy loss, we train E-IIET to effectively capture the knowledge embedded within the full IIET's iterative refinement process. The final training objective for this continuous pre-training stage is thus:
\begin{equation}
\mathcal{L}_{\text{E-IIET}} = \mathcal{L}_\text{CE} + \alpha\mathcal{L}_\text{MSE} + \beta\mathcal{L}_\text{KL}
\end{equation}

\subsection{Experiments and Results}

\paragraph{Setups.} To train efficient IIET variants, we sample one-third of the original pre-training tokens (see Appendix~\ref{sec:distill_detail} for detailed training settings). For performance comparison against E-IIET, we also prepare two key baselines: a \textit{Lower Bound} variant, which omits all iterative corrections, and a distilled version of PCformer. All models are trained following the method outlined in Section~\ref{sec:iiad_method}.

\paragraph{Main Results.} Table~\ref{tab:main_results} presents the main results for IIAD. As a baseline, directly distilling PCformer into a standard Euler architecture (namely \textit{Distil PCformer}) leads to substantial performance degradation, highlighting the importance of the sophisticated numerical solvers employed by higher-order methods to achieve their accuracy. In contrast, E-IIET, compared to the full IIET model, retains the vast majority of its performance while reducing the average iterative correction overhead by about 55\%. Importantly, even the \textit{Lower Bound} efficient IIET variant achieves performance on par with PCformer, demonstrating IIET's strength in balancing efficiency with strong performance. 

\begin{table}
\setlength{\tabcolsep}{1.2pt}
\footnotesize
\centering
\begin{tabular}{@{}l|ccc|ccc@{}}
\toprule
\multirow{2}{*}{\textbf{Model}} & 
\multicolumn{3}{c}{\textbf{340M}} & 
\multicolumn{3}{c}{\textbf{740M}} \\
\cmidrule(lr){2-4} \cmidrule(lr){5-7}
& \multicolumn{1}{c}{\textbf{Spd.}} & \multicolumn{1}{c}{\textbf{FLOPs}} & \multicolumn{1}{c}{\textbf{VRAM}} & \multicolumn{1}{c}{\textbf{Spd.}} & \multicolumn{1}{c}{\textbf{FLOPs}} & \multicolumn{1}{c}{\textbf{VRAM}} \\

\midrule
Transformer++	& 49.97 & 0.38 & 1.37 & 48.91 & 0.80 & 2.80 \\
PCformer	& 14.14 & 1.06 & 1.41 & 14.38 & 2.30 & 2.86 \\
IIET	& 11.07 & 1.40 & 1.42 & 10.95 & 3.05 & 2.89 \\


\midrule
Lower Bound & 42.66 & 0.38 & 1.37 & 42.03 & 0.80 & 2.80 \\
E-IIET & 25.95 & 0.60 & 1.38 & 22.12 & 1.52 & 2.83 \\

\bottomrule
\end{tabular}
\caption{A comparison of inference speed (tokens per second), FLOPs (T) and VRAM (GB) for baseline models, PCformer, and efficient IIET variants.}
\label{tab:inference_speed}
\vspace{-3mm}
\end{table}

\paragraph{Inference Efficiency.} 
We analyze the inference speed, FLOPs and VRAM usage of our main models. As Table~\ref{tab:inference_speed} indicates, E-IIET achieves over a 2x speedup compared to full IIET, while largely maintaining IIET's performance advantage (E-IIET vs. full IIET scores: 48.9/48.9 for 340M and 51.3/51.9 for 740M model). However, due to the FLOPs incurred by its remaining iteration steps, E-IIET still exhibits nearly twice the inference latency of Transformer++. A key characteristic of these efficient IIET variants is the inverse relationship between performance and efficiency: fewer iterations lead to lower performance but higher efficiency. Notably, Table~\ref{tab:inference_speed} shows that the maximum efficiency attained by these variants (i.e, \textit{Lower Bound}) is close to that of the Transformer++, with their average performance surpassing it by 1.6 points. This adaptability makes E-IIET a flexible solution for practical deployment, as users can select the iteration steps based on their resource constraints (e.g., reducing iterations to maximize inference speed).


\section{Related Work}

\paragraph{Ordinary Differential Equations in Deep Learning} The conceptual link between Ordinary Differential Equations (ODEs) and residual networks, first established by \citet{weinan2017proposal}, has catalyzed the development of numerous ODE-inspired neural architectures. In computer vision, this perspective give rise to models such as PolyNet~\cite{zhang2017polynet}, FractalNet~\cite{larsson2016fractalnet}, Multi-stepNet~\cite{lu2018beyond}, and Momentum Residual Networks~\cite{sander2021momentum}. The ODE viewpoint has also been highly influential in generative modeling, particularly for diffusion models~\cite{ho2020denoising}. For example, \citet{liu2022pseudo} re-framed Denoising Diffusion Probabilistic Models (DDPMs) as a process of solving differential equations on manifolds, introducing a pseudo linear multi-step method to improve performance. Building on this, DPM-Solver~\cite{lu2022dpm,lu2022dpm2} significantly accelerated the sampling process by employing exact ODE solutions and higher-order numerical methods. More recently, ODE principles have been leveraged to enhance Transformer architectures for sequence modeling and generation~\cite{lu2019understanding,wang2019learning}. \citet{dutta2021redesigning} redesigned the Transformer as a more efficient multi-particle dynamic system, while \citet{tong2025neural} proposed high-order methods to mitigate error accumulation in first-order ODE blocks. Further advancing this line of work, PCformer~\cite{li2024predictor} introduced a predictor-corrector framework to boost performance, and they first show the potential of such numerical design in large language model literature. In our work, we build upon this foundation by employing the implicit Euler method, aiming to enhance language modeling performance while achieving a superior trade-off between accuracy and computational efficiency.

\paragraph{Implicit ODE Method} Existing approaches that utilize implicit ODE solvers can be broadly categorized into two paradigms. The first, exemplified by seminal works like Neural ODEs~\cite{chen2018neural,zhang2021continuous} and Deep Equilibrium Models (DEQs)~\cite{bai2019deep}, represents a significant shift towards implicit deep learning. Neural ODEs model the network as a continuous transformation by using a neural network to parameterize the state's derivative, which is solved with a numerical ODE solver. In contrast, DEQs define network layers implicitly through an equilibrium point, denoted as $z^*=f(z^*,x)$, which is found iteratively, with gradients computed via implicit differentiation. While subsequent research has explored these models for applications like time-series forecasting, their computational cost remains a significant barrier for large-scale language modeling; for instance, a DEQ model can require over five times the computation to match the performance of a standard Transformer-XL model (240M). More recently, the Neural ODE Transformer~\cite{tong2025neural} has achieved performance surpassing that of the vanilla Transformer on language modeling tasks. Another class of methods utilizes advanced implicit numerical solvers to optimize mainstream model architectures~\cite{li2024predictor,li2020implicit,shen2020implicit,kim2024bert}. For instance, IE-Skips~\cite{li2020implicit} modifies the original skip connection in ResNet for robustness. Similarly, IM-BERT~\cite{kim2024bert} introduced the use of the implicit Euler method to enhance the adversarial robustness of BERT. Our proposed IIET distinctively focuses on autoregressive generation, and the core formulation is quite different, that we adopted an iterative refinement schema began from a fixed point. To our knowledge we are the first to apply iterative implict Euler into LLMs paradigm.

\section{Conclusions}


We propose a novel Transformer architecture, the Iterative Implicit Euler Transformer (IIET), designed for enhanced language modeling performance. IIET leverages the iterative implicit Euler method, providing substantial improvements over vanilla Transformers with a simplified architecture compared to PCformer. Furthermore, we introduce an inference acceleration technique for IIET, which uses self-distillation to prune the iterative process, allowing users to adjust inference efficiency based on their budget.


\section*{Limitations}
Although the IIAD method is designed to produce efficient IIET variants for inference, the IIAD process itself introduces notable computational overhead during its application. Future research will focus on integrating the determination of layer-specific iteration requirements directly into the pre-training stage. This could facilitate the direct training of inherently efficient IIET models, potentially bypassing a separate, resource-intensive distillation phase.

Beyond optimizing IIET's per-token efficiency, we also identify a promising avenue for broader application. Current large reasoning models often achieve high performance by generating substantially more tokens than are present in the final answer, leading to significant inference latency. IIET, on the contrary, enhances per token representational power through depth-wise iterative refinement, albeit at an increased per-token computational cost. We hypothesize that this trade-off could be ultimately advantageous in multi-step reasoning tasks: IIET's more precise computation per token might enable it to generate complete and correct answers in fewer overall autoregressive steps, thereby reducing the total token count and potentially overall latency. Validating this hypothesis, however, necessitates training and evaluating IIET at larger model and data scales, which remains a key direction for future investigation.

\section*{Acknowledgments}
This work was supported in part by the National Science Foundation of China (Nos. 62276056 and U24A20334), the Yunnan Fundamental Research Projects (No.202401BC070021), the Yunnan Science and Technology Major Project (No. 202502AD080014), and the Program of Introducing Talents of Discipline to Universities, Plan 111 (No.B16009).

\bibliography{custom}

\begin{thebibliography}{47}
\providecommand{\natexlab}[1]{#1}

\bibitem[{Bai et~al.(2019)Bai, Kolter, and Koltun}]{bai2019deep}
Shaojie Bai, J~Zico Kolter, and Vladlen Koltun. 2019.
\newblock Deep equilibrium models.
\newblock \emph{Advances in neural information processing systems}, 32.

\bibitem[{Bisk et~al.(2020)Bisk, Zellers, Gao, Choi et~al.}]{bisk2020piqa}
Yonatan Bisk, Rowan Zellers, Jianfeng Gao, Yejin Choi, and 1 others. 2020.
\newblock Piqa: Reasoning about physical commonsense in natural language.
\newblock In \emph{Proceedings of the AAAI conference on artificial intelligence}, volume~34, pages 7432--7439.

\bibitem[{Chen et~al.(2018)Chen, Rubanova, Bettencourt, and Duvenaud}]{chen2018neural}
Ricky~TQ Chen, Yulia Rubanova, Jesse Bettencourt, and David~K Duvenaud. 2018.
\newblock Neural ordinary differential equations.
\newblock \emph{Advances in neural information processing systems}, 31.

\bibitem[{Chen et~al.(2024)Chen, Hu, and Zhang}]{chen2024compressing}
Xiaodong Chen, Yuxuan Hu, and Jing Zhang. 2024.
\newblock Compressing large language models by streamlining the unimportant layer.
\newblock \emph{arXiv preprint arXiv:2403.19135}.

\bibitem[{Clark et~al.(2018)Clark, Cowhey, Etzioni, Khot, Sabharwal, Schoenick, and Tafjord}]{clark2018think}
Peter Clark, Isaac Cowhey, Oren Etzioni, Tushar Khot, Ashish Sabharwal, Carissa Schoenick, and Oyvind Tafjord. 2018.
\newblock Think you have solved question answering? try arc, the ai2 reasoning challenge.
\newblock \emph{arXiv preprint arXiv:1803.05457}.

\bibitem[{Dutta et~al.(2021)Dutta, Gautam, Chakrabarti, and Chakraborty}]{dutta2021redesigning}
Subhabrata Dutta, Tanya Gautam, Soumen Chakrabarti, and Tanmoy Chakraborty. 2021.
\newblock Redesigning the transformer architecture with insights from multi-particle dynamical systems.
\newblock \emph{Advances in Neural Information Processing Systems}, 34:5531--5544.

\bibitem[{Gao et~al.(2021)Gao, Tow, Biderman, Black, DiPofi, Foster, Golding, Hsu, McDonell, Muennighoff et~al.}]{gao2021framework}
Leo Gao, Jonathan Tow, Stella Biderman, Sid Black, Anthony DiPofi, Charles Foster, Laurence Golding, Jeffrey Hsu, Kyle McDonell, Niklas Muennighoff, and 1 others. 2021.
\newblock A framework for few-shot language model evaluation.
\newblock \emph{Version v0. 0.1. Sept}, 10:8--9.

\bibitem[{Gu and Dao(2023)}]{Gu2023Mamba}
Albert Gu and Tri Dao. 2023.
\newblock Mamba: Linear-time sequence modeling with selective state spaces.
\newblock \emph{arXiv preprint arXiv:2312.00752}.

\bibitem[{Hinton(2015)}]{hinton2015distilling}
Geoffrey Hinton. 2015.
\newblock Distilling the knowledge in a neural network.
\newblock \emph{arXiv preprint arXiv:1503.02531}.

\bibitem[{Ho et~al.(2020)Ho, Jain, and Abbeel}]{ho2020denoising}
Jonathan Ho, Ajay Jain, and Pieter Abbeel. 2020.
\newblock Denoising diffusion probabilistic models.
\newblock \emph{Advances in neural information processing systems}, 33:6840--6851.

\bibitem[{Hu et~al.(2024)Hu, Tu, Han, He, Cui, Long, Zheng, Fang, Huang, Zhao et~al.}]{hu2024minicpm}
Shengding Hu, Yuge Tu, Xu~Han, Chaoqun He, Ganqu Cui, Xiang Long, Zhi Zheng, Yewei Fang, Yuxiang Huang, Weilin Zhao, and 1 others. 2024.
\newblock Minicpm: Unveiling the potential of small language models with scalable training strategies.
\newblock \emph{arXiv preprint arXiv:2404.06395}.

\bibitem[{Kim et~al.(2024)Kim, Park, and Kim}]{kim2024bert}
Mihyeon Kim, Juhyoung Park, and Youngbin Kim. 2024.
\newblock Im-bert: Enhancing robustness of bert through the implicit euler method.
\newblock In \emph{Proceedings of the 2024 Conference on Empirical Methods in Natural Language Processing}, pages 16217--16229.

\bibitem[{Kim and Rush(2016)}]{kim2016sequence}
Yoon Kim and Alexander~M Rush. 2016.
\newblock Sequence-level knowledge distillation.
\newblock \emph{arXiv preprint arXiv:1606.07947}.

\bibitem[{Larsson et~al.(2016)Larsson, Maire, and Shakhnarovich}]{larsson2016fractalnet}
Gustav Larsson, Michael Maire, and Gregory Shakhnarovich. 2016.
\newblock Fractalnet: Ultra-deep neural networks without residuals.
\newblock \emph{arXiv preprint arXiv:1605.07648}.

\bibitem[{LeVeque(2007)}]{finite2007siam}
Randall~J. LeVeque. 2007.
\newblock \emph{Finite difference methods for ordinary and partial differential equations - steady-state and time-dependent problems}.
\newblock {SIAM}.

\bibitem[{Li et~al.(2022)Li, Du, Zhou, Jing, Zhou, Zeng, Xiao, Zhu, Liu, and Zhang}]{li2022ode}
Bei Li, Quan Du, Tao Zhou, Yi~Jing, Shuhan Zhou, Xin Zeng, Tong Xiao, JingBo Zhu, Xuebo Liu, and Min Zhang. 2022.
\newblock Ode transformer: An ordinary differential equation-inspired model for sequence generation.
\newblock \emph{arXiv preprint arXiv:2203.09176}.

\bibitem[{Li et~al.(2024)Li, Zheng, Wang, Liu, Guo, Guo, Tan, Xiao, Zhu, Wang et~al.}]{li2024predictor}
Bei Li, Tong Zheng, Rui Wang, Jiahao Liu, Qingyan Guo, Junliang Guo, Xu~Tan, Tong Xiao, Jingbo Zhu, Jingang Wang, and 1 others. 2024.
\newblock Predictor-corrector enhanced transformers with exponential moving average coefficient learning.
\newblock \emph{arXiv preprint arXiv:2411.03042}.

\bibitem[{Li et~al.(2020)Li, He, and Lin}]{li2020implicit}
Mingjie Li, Lingshen He, and Zhouchen Lin. 2020.
\newblock Implicit euler skip connections: Enhancing adversarial robustness via numerical stability.
\newblock In \emph{International Conference on Machine Learning}, pages 5874--5883. PMLR.

\bibitem[{Liu et~al.(2024)Liu, Feng, Xue, Wang, Wu, Lu, Zhao, Deng, Zhang, Ruan et~al.}]{liu2024deepseek}
Aixin Liu, Bei Feng, Bing Xue, Bingxuan Wang, Bochao Wu, Chengda Lu, Chenggang Zhao, Chengqi Deng, Chenyu Zhang, Chong Ruan, and 1 others. 2024.
\newblock Deepseek-v3 technical report.
\newblock \emph{arXiv preprint arXiv:2412.19437}.

\bibitem[{Liu et~al.(2022)Liu, Ren, Lin, and Zhao}]{liu2022pseudo}
Luping Liu, Yi~Ren, Zhijie Lin, and Zhou Zhao. 2022.
\newblock Pseudo numerical methods for diffusion models on manifolds.
\newblock \emph{arXiv preprint arXiv:2202.09778}.

\bibitem[{Lu et~al.(2022{\natexlab{a}})Lu, Zhou, Bao, Chen, Li, and Zhu}]{lu2022dpm}
Cheng Lu, Yuhao Zhou, Fan Bao, Jianfei Chen, Chongxuan Li, and Jun Zhu. 2022{\natexlab{a}}.
\newblock Dpm-solver: A fast ode solver for diffusion probabilistic model sampling in around 10 steps.
\newblock \emph{Advances in Neural Information Processing Systems}, 35:5775--5787.

\bibitem[{Lu et~al.(2022{\natexlab{b}})Lu, Zhou, Bao, Chen, Li, and Zhu}]{lu2022dpm2}
Cheng Lu, Yuhao Zhou, Fan Bao, Jianfei Chen, Chongxuan Li, and Jun Zhu. 2022{\natexlab{b}}.
\newblock Dpm-solver++: Fast solver for guided sampling of diffusion probabilistic models.
\newblock \emph{arXiv preprint arXiv:2211.01095}.

\bibitem[{Lu et~al.(2019)Lu, Li, He, Sun, Dong, Qin, Wang, and Liu}]{lu2019understanding}
Yiping Lu, Zhuohan Li, Di~He, Zhiqing Sun, Bin Dong, Tao Qin, Liwei Wang, and Tie-Yan Liu. 2019.
\newblock Understanding and improving transformer from a multi-particle dynamic system point of view.
\newblock \emph{arXiv preprint arXiv:1906.02762}.

\bibitem[{Lu et~al.(2018)Lu, Zhong, Li, and Dong}]{lu2018beyond}
Yiping Lu, Aoxiao Zhong, Quanzheng Li, and Bin Dong. 2018.
\newblock Beyond finite layer neural networks: Bridging deep architectures and numerical differential equations.
\newblock In \emph{International Conference on Machine Learning}, pages 3276--3285. PMLR.

\bibitem[{Men et~al.(2024)Men, Xu, Zhang, Wang, Lin, Lu, Han, and Chen}]{men2024shortgpt}
Xin Men, Mingyu Xu, Qingyu Zhang, Bingning Wang, Hongyu Lin, Yaojie Lu, Xianpei Han, and Weipeng Chen. 2024.
\newblock Shortgpt: Layers in large language models are more redundant than you expect.
\newblock \emph{arXiv preprint arXiv:2403.03853}.

\bibitem[{Merity et~al.(2016)Merity, Xiong, Bradbury, and Socher}]{merity2016pointer}
Stephen Merity, Caiming Xiong, James Bradbury, and Richard Socher. 2016.
\newblock Pointer sentinel mixture models.
\newblock \emph{arXiv preprint arXiv:1609.07843}.

\bibitem[{Paperno et~al.(2016)Paperno, Kruszewski, Lazaridou, Pham, Bernardi, Pezzelle, Baroni, Boleda, and Fern{\'a}ndez}]{paperno2016lambada}
Denis Paperno, Germ{\'a}n Kruszewski, Angeliki Lazaridou, Quan~Ngoc Pham, Raffaella Bernardi, Sandro Pezzelle, Marco Baroni, Gemma Boleda, and Raquel Fern{\'a}ndez. 2016.
\newblock The lambada dataset: Word prediction requiring a broad discourse context.
\newblock \emph{arXiv preprint arXiv:1606.06031}.

\bibitem[{Rhoades(1976)}]{rhoades1976comments}
BE~Rhoades. 1976.
\newblock Comments on two fixed point iteration methods.
\newblock \emph{Journal of Mathematical Analysis and Applications}, 56(3):741--750.

\bibitem[{Sakaguchi et~al.(2021)Sakaguchi, Bras, Bhagavatula, and Choi}]{sakaguchi2021winogrande}
Keisuke Sakaguchi, Ronan~Le Bras, Chandra Bhagavatula, and Yejin Choi. 2021.
\newblock Winogrande: An adversarial winograd schema challenge at scale.
\newblock \emph{Communications of the ACM}, 64(9):99--106.

\bibitem[{Sander et~al.(2021)Sander, Ablin, Blondel, and Peyr{\'e}}]{sander2021momentum}
Michael~E Sander, Pierre Ablin, Mathieu Blondel, and Gabriel Peyr{\'e}. 2021.
\newblock Momentum residual neural networks.
\newblock In \emph{International Conference on Machine Learning}, pages 9276--9287. PMLR.

\bibitem[{Shazeer(2020)}]{shazeer2020glu}
Noam Shazeer. 2020.
\newblock Glu variants improve transformer.
\newblock \emph{arXiv preprint arXiv:2002.05202}.

\bibitem[{Shen et~al.(2020)Shen, Li, Yu, Xia, and Yang}]{shen2020implicit}
Jiawei Shen, Zhuoyan Li, Lei Yu, Gui-Song Xia, and Wen Yang. 2020.
\newblock Implicit euler ode networks for single-image dehazing.
\newblock In \emph{Proceedings of the IEEE/CVF Conference on Computer Vision and Pattern Recognition Workshops}, pages 218--219.

\bibitem[{Soboleva et~al.(2023)Soboleva, Al-Khateeb, Myers, Steeves, Hestness, and Dey}]{soboleva2023slimpajama}
Daria Soboleva, Faisal Al-Khateeb, Robert Myers, Jacob~R Steeves, Joel Hestness, and Nolan Dey. 2023.
\newblock Slimpajama: A 627b token cleaned and deduplicated version of redpajama.

\bibitem[{Su et~al.(2024)Su, Ahmed, Lu, Pan, Bo, and Liu}]{su2024roformer}
Jianlin Su, Murtadha Ahmed, Yu~Lu, Shengfeng Pan, Wen Bo, and Yunfeng Liu. 2024.
\newblock Roformer: Enhanced transformer with rotary position embedding.
\newblock \emph{Neurocomputing}, 568:127063.

\bibitem[{Tong et~al.(2025)Tong, Nguyen-Tang, Lee, Nguyen, Tran, Hall, Kang, and Choi}]{tong2025neural}
Anh Tong, Thanh Nguyen-Tang, Dongeun Lee, Duc Nguyen, Toan Tran, David Leo~Wright Hall, Cheongwoong Kang, and Jassik Choi. 2025.
\newblock Neural ode transformers: Analyzing internal dynamics and adaptive fine-tuning.
\newblock In \emph{ICT.R.2025 Poster}.
\newblock Unpublished.

\bibitem[{Touvron et~al.(2023{\natexlab{a}})Touvron, Lavril, Izacard, Martinet, Lachaux, Lacroix, Rozi{\`e}re, Goyal, Hambro, Azhar et~al.}]{touvron2023llama}
Hugo Touvron, Thibaut Lavril, Gautier Izacard, Xavier Martinet, Marie-Anne Lachaux, Timoth{\'e}e Lacroix, Baptiste Rozi{\`e}re, Naman Goyal, Eric Hambro, Faisal Azhar, and 1 others. 2023{\natexlab{a}}.
\newblock Llama: Open and efficient foundation language models.
\newblock \emph{arXiv preprint arXiv:2302.13971}.

\bibitem[{Touvron et~al.(2023{\natexlab{b}})Touvron, Martin, Stone, Albert, Almahairi, Babaei, Bashlykov, Batra, Bhargava, Bhosale et~al.}]{touvron2023llama2}
Hugo Touvron, Louis Martin, Kevin Stone, Peter Albert, Amjad Almahairi, Yasmine Babaei, Nikolay Bashlykov, Soumya Batra, Prajjwal Bhargava, Shruti Bhosale, and 1 others. 2023{\natexlab{b}}.
\newblock Llama 2: Open foundation and fine-tuned chat models.
\newblock \emph{arXiv preprint arXiv:2307.09288}.

\bibitem[{Vaswani(2017)}]{vaswani2017attention}
A~Vaswani. 2017.
\newblock Attention is all you need.
\newblock \emph{Advances in Neural Information Processing Systems}.

\bibitem[{Wang et~al.(2019)Wang, Li, Xiao, Zhu, Li, Wong, and Chao}]{wang2019learning}
Qiang Wang, Bei Li, Tong Xiao, Jingbo Zhu, Changliang Li, Derek~F Wong, and Lidia~S Chao. 2019.
\newblock Learning deep transformer models for machine translation.
\newblock \emph{arXiv preprint arXiv:1906.01787}.

\bibitem[{Weinan(2017)}]{weinan2017proposal}
Ee~Weinan. 2017.
\newblock A proposal on machine learning via dynamical systems.
\newblock \emph{Communications in Mathematics and Statistics}, 1(5):1--11.

\bibitem[{Welbl et~al.(2017)Welbl, Liu, and Gardner}]{welbl2017crowdsourcing}
Johannes Welbl, Nelson~F Liu, and Matt Gardner. 2017.
\newblock Crowdsourcing multiple choice science questions.
\newblock \emph{arXiv preprint arXiv:1707.06209}.

\bibitem[{Zellers et~al.(2019)Zellers, Holtzman, Bisk, Farhadi, and Choi}]{zellers2019hellaswag}
Rowan Zellers, Ari Holtzman, Yonatan Bisk, Ali Farhadi, and Yejin Choi. 2019.
\newblock Hellaswag: Can a machine really finish your sentence?
\newblock \emph{arXiv preprint arXiv:1905.07830}.

\bibitem[{Zhang and Sennrich(2019)}]{zhang2019root}
Biao Zhang and Rico Sennrich. 2019.
\newblock Root mean square layer normalization.
\newblock \emph{Advances in Neural Information Processing Systems}, 32.

\bibitem[{Zhang et~al.(2021)Zhang, Zhang, Kong, Wei, and Jiang}]{zhang2021continuous}
Jing Zhang, Peng Zhang, Baiwen Kong, Junqiu Wei, and Xin Jiang. 2021.
\newblock Continuous self-attention models with neural ode networks.
\newblock In \emph{Proceedings of the AAAI conference on artificial intelligence}, volume~35, pages 14393--14401.

\bibitem[{Zhang et~al.(2017)Zhang, Li, Change~Loy, and Lin}]{zhang2017polynet}
Xingcheng Zhang, Zhizhong Li, Chen Change~Loy, and Dahua Lin. 2017.
\newblock Polynet: A pursuit of structural diversity in very deep networks.
\newblock In \emph{Proceedings of the IEEE conference on computer vision and pattern recognition}, pages 718--726.

\bibitem[{Zhao et~al.(2024)Zhao, Bai, Rao, Zhou, and Lu}]{zhao2024unipc}
Wenliang Zhao, Lujia Bai, Yongming Rao, Jie Zhou, and Jiwen Lu. 2024.
\newblock Unipc: A unified predictor-corrector framework for fast sampling of diffusion models.
\newblock \emph{Advances in Neural Information Processing Systems}, 36.

\bibitem[{Zheng et~al.(2024)Zheng, Lu, Chen, and Zhu}]{zheng2024dpm}
Kaiwen Zheng, Cheng Lu, Jianfei Chen, and Jun Zhu. 2024.
\newblock Dpm-solver-v3: Improved diffusion ode solver with empirical model statistics.
\newblock \emph{Advances in Neural Information Processing Systems}, 36.

\end{thebibliography}

\newpage

\appendix

\section{Training Settings}
\subsection{Pre-training Phase}
\label{sec:pretrain_detail}
To evaluate IIET's performance across different model sizes, we train models from scratch at three parameter scales: 340M, 740M, and 1.3B. For all training runs, we employ the AdamW optimizer with a maximum learning rate of 3e-4. A batch size of 0.5M tokens is used for the 340M model, while 1M tokens are used for the 740M and 1.3B models. We apply a cosine learning rate schedule to all model scales, which includes a 0.01 warmup ratio, 0.01 weight decay, and gradient clipping at 1.0. Furthermore, a smaller 55M parameter IIET variant is trained to determine the optimal iteration count, $r$. The complete hyperparameter details for the pre-training phase are provided in Table~\ref{tab:hyperparameters}.

\begin{table*}[htp!]
\centering
    \centering
    \begin{tabular}{lllll}
    \toprule
    \textbf{Hyperparameters} & \textbf{55M} & \textbf{340M} & \textbf{740M} & \textbf{1.3B} \\
    \midrule
    model\_type & llama & llama & llama & llama \\
    hidden\_act & silu & silu & silu & silu \\
    initializer\_range & 0.02 & 0.02 & 0.02 & 0.02 \\
    hidden\_size & 512 & 1024 & 1536 & 2048 \\
    intermediate\_size & 1408 & 2816 & 4224 & 5504 \\
    max\_position\_embeddings & 2048 & 2048 & 2048 & 2048 \\
    num\_attention\_heads & 4 & 8 & 8 & 16 \\
    num\_hidden\_layers & 12 & 24 & 24 & 24 \\
    num\_key\_value\_heads & 4 & 8 & 8 & 16 \\
    pretraining\_tp & 1 & 1 & 1 & 1 \\
    rms\_norm\_eps & $1.00 \times 10^{-6}$ & $1.00 \times 10^{-6}$ & $1.00 \times 10^{-6}$ & $1.00 \times 10^{-6}$ \\
    tie\_word\_embeddings & True & True & True & False \\
    torch\_dtype & float16 & float16 & float16 & float16 \\
    vocab\_size & 32000 & 32000 & 32000 & 32000 \\
    \midrule
    training\_len & 2048 & 2048 & 2048 & 2048 \\
    total\_batch\_size & 128 & 256 & 512 & 512 \\
    learning\_rate & 0.0004 & 0.0003 & 0.0003 & 0.0002 \\
    max\_steps & 5000 & 30000 & 30000 & 100000 \\
    warm\_up & 0.05 & 0.05 & 0.01 & 0.01 \\
    \bottomrule
    \end{tabular}
    \caption{The key hyperparameters for both the model architecture and the training process.}
    \vspace{-3mm}
    \label{tab:hyperparameters}
\end{table*}

\subsection{Iteration Influence-Aware Distillation Phase}
\label{sec:distill_detail}

To train efficient IIET variants, we sample one-third of the total pre-training tokens for each configuration (e.g., 5 billion tokens for 340M models, 10 billion tokens for 740M models and 30 billion tokens for 1.3B models). Users can customize the corrective iteration process for these variants based on their computational budget. In this study, we focus on two main types of efficient IIETs: a `lower bound' configuration that removes all iterative steps, and E-IIET, which utilizes a threshold for iteration selection. For training, all efficient IIET variants use the full IIET as a teacher model and are trained with the fine-grained supervision method detailed in Section~\ref{sec:iiad_method}. We apply a cosine decay learning rate schedule with an initial value of 2e-4, while other pre-training hyperparameters are kept consistent. Furthermore, for comparison purposes, we train Distil PCformer, a self-distilled version of PCformer using the same methodology. To ensure a fair comparison, we use the same evaluation dataset and metrics as described in Section~\ref{sec:exp_results}.

\subsection{Two-Stage Training}
\label{sec:two_stage_detail}
We conduct the two-stage training experiment focusing on a model with a 740 million parameter scale. The specific model configurations are provided in Table~\ref{tab:hyperparameters}. In the initial stage, we train a standard Transformer architecture with a linear learning rate warmup followed by a constant rate. In the second stage, we adapt the model to the IIET structure with $r=3$. The detailed hyperparameters for both training phases are summarized in Table~\ref{tab:two_stage_hyperparameters}.

\begin{table}[t]
\centering
    \centering
    \small
    \begin{tabular}{lc}
    \toprule
    \textbf{Hyperparameters} & \textbf{Value} \\
    \midrule
    \vspace{-1.2em} \\
    \multicolumn{2}{l}{\textit{Foundational Pretraining}} \\
    \vspace{-1.2em} \\
    \midrule
    batch\_size & 512 \\
    learning\_rate & 3e-4 \\
    lr\_scheduler\_type & constant\_with\_warmup \\
    warmup\_steps & 300 \\
    training\_len & 2048 \\
    total\_steps & 22,000 \\
    \midrule
    \vspace{-1.2em} \\
    \multicolumn{2}{l}{\textit{Architectural Transition}} \\
    \vspace{-1.2em} \\
    \midrule
    batch\_size & 512 \\
    learning\_rate & 3e-4 \\
    lr\_scheduler\_type & cosine \\
    warmup\_steps & 0 \\
    training\_len & 2048 \\
    total\_steps & 8,000 \\
    \bottomrule
    \end{tabular}
    \caption{Hyperparameter configuration for two-stage training.}
    \vspace{-3mm}
    \label{tab:two_stage_hyperparameters}
\end{table}

\section{Experimental Results on Other Tasks}
\label{sec:results_on_other_tasks}
To validate the generalizability of our proposed IIET architecture, we present experimental results on machine translation and summarization. For a fair and direct comparison, our experimental setup strictly follows the one established in PCformer~\cite{li2024predictor}. Specifically, we train IIET in a standard big configuration (6-layer encoder and 6-layer decoder), with three iterations of our iterative implicit Euler method applied to the encoder, which is consistent with the methodology in our main experiments. Results for all baseline models, including PCformer, are taken directly from the original publication. For the OPUS benchmark, we report the average score of the X-En and En-X translation directions.

\begin{table}[t]
    \centering
    \scalebox{0.71}{  
    \begin{tabular}{l cccc}
    \toprule
    \textbf{Model} & {\thead{En-De \\ BLEU}} & {\thead{En-Fr \\ BLEU}} & {\thead{OPUS \\ SacreBLEU}} & {\thead{Summarization \\ Rouge-1/2/L}} \\
    \midrule
    Transformer++ & 29.2 & 42.9 & 30.8 (34.0/27.6) & 40.5/17.7/37.3 \\
    PCformer & 30.9 & 43.9 & 32.6 (36.0/29.1) & 42.0/19.0/38.7 \\
    IIET & 30.7 & 43.8 & 32.1 (35.7/28.6) &	42.0/19.0/38.7 \\
    \bottomrule
    \end{tabular}}
    \caption{Performance comparison of IIET against baseline models on machine translation (WMT14) and summarization tasks.}
    \vspace{-3mm}
    \label{tab:other_tasks}
\end{table}

The results in Table~\ref{tab:other_tasks} confirm that IIET achieves performance competitive with the state-of-the-art PCformer model while substantially outperforming the strong Transformer++ baseline. However, we posit that IIET's primary practical advantage lies in its superior adaptability for low-FLOPs deployment in resource-constrained environments. This adaptability contrasts with models like PCformer, where the inherent predictor-corrector discrepancy can hinder effective compression.
Furthermore, our training logs reveal an increase in validation perplexity during the final training stages, which we attribute to overfitting. This observation suggests that even stronger performance may be achievable through simple hyperparameter tuning, such as increasing the dropout rate. We leave this exploration for future work.

\begin{table}[t]
    \centering
    \scalebox{0.57}{  
    \begin{tabular}{l cccccccc}
    \toprule
    \textbf{Model} & {\textbf{Iter.}} & {\textbf{LMB.}} & {\textbf{PiQA}} & {\textbf{Hella.}} & {\textbf{SCIQ}} & {\textbf{ARC-c}} & {\textbf{Wino.}} & {\textbf{Avg.}} \\
    \midrule
    Transformer++ & - & 28.9 & 64.3 & 34.3 & 76.0 & 23.6 & 51.9 & 46.5 \\
    PCformer & - & 33.1 & 64.9 & 36.3 & 77.5 & 24.7 & 53.3 & 48.3 \\
    \midrule
    IIET & 0 & 32.4 & 65.1 & 34.8 & 78.3 & 23.5 & 50.4 & 47.4 \\
    IIET & 1 & 34.4 & 64.7 & 36.1 & 76.3 & 23.3 & 50.1 & 47.5 \\
    IIET & 2 & 34.6 & 65.0 & 36.8 & 77.2 & 24.2 & 51.9 & 48.3 \\
    IIET & 3 & 37.1 & 65.2 & 36.9 & 79.4 & 23.9 & 51.0 & 48.9 \\
    IIET & 4 & 36.8 & 64.3 & 37.3 & 78.1 & 22.8 & 53.8 & 48.8 \\
    IIET & 5 & 36.3 & 64.5 & 37.3 & 78.6 & 23.4 & 53.2 & 48.9 \\
    IIET & 6 & 35.8 & 64.7 & 37.4 & 79.0 & 24.1 & 52.9 & 49.0 \\
    IIET & 7 & 35.5 & 65.2 & 37.0 & 79.2 & 23.3 & 53.0 & 48.9 \\
    IIET & 8 & 35.2 & 65.6 & 36.5 & 79.4 & 22.6 & 51.1 & 48.4 \\
    \bottomrule
    \end{tabular}}
    \caption{Performance comparison of IIET with varying iteration steps at 340 million parameters.}
    \vspace{-3mm}
    \label{tab:analysis_iter_count_perf}
\end{table}
\section{IIET with Varying Iteration Steps}
\label{sec:analysis_iter_step_detail}

We evaluated the downstream task performance of our 340M model across iteration steps $r=0$ to $r=8$, as detailed in Section~\ref{sec:ablation_r}. Table~\ref{tab:analysis_iter_count_perf} shows that as the number of iterations increases, IIET's performance on downstream tasks initially improves progressively before these gains begin to plateau. Although performance slightly degrades at $r=8$, IIET still surpasses both Transformer++ and PCformer. Notably, with $r=2$ iterations, IIET achieves performance comparable to PCformer with its per-block forward pass count is also similar to PCformer's. This demonstrates that our proposed iterative implicit Euler (IIET) architecture, despite its simpler design, offers representation refinement capabilities that are close to those of higher-order methods. Finally, using identical training data, IIET exhibited superior data-fitting ability over the other models, as indicated by its perplexity scores.

\section{Comparison with Compressed Vanilla Models}
\label{sec:comparison_with_compressed_models}

To ensure a fair comparison based on model size, we compressed the 1.3B Transformer++ model (trained on 16B tokens) using pruning and quantization. These compressed baselines are further compared with the 340M IIET to demonstrate the structural advantages of IIET.

\paragraph{Quantization} We employ the AutoGPTQ library\footnote{\url{https://github.com/AutoGPTQ/AutoGPTQ}} to perform 4-bit quantization on the 1.3B Transformer++. This process produces Trans quant, a model with a parameter storage footprint comparable to the 340M IIET model in bfloat16 format. The quantization is calibrated on a set of 512 samples from the C4 dataset. We apply the quantization using a group size of 128, a damping percentage of 0.01, and the standard activation order. Table~\ref{tab:comparison_compress} shows that Transformer++ (4-bit) experiences a slight performance degradation compared to 1.3B Transformer++ and also underperforms the 340M IIET model (47.8 vs 48.9). The results suggest that IIET offers structural advantages beyond what post-hoc compression can achieve. Moreover, we do not apply quantization to the activations, which is in line with common practice. As a result, while 4-bit quantization reduces the parameter storage footprint, it does not lead to a significant reduction in FLOPs, since activations are still processed at 16-bit precision.

\begin{figure*}[t]
  \centering
  \includegraphics[width=\textwidth]{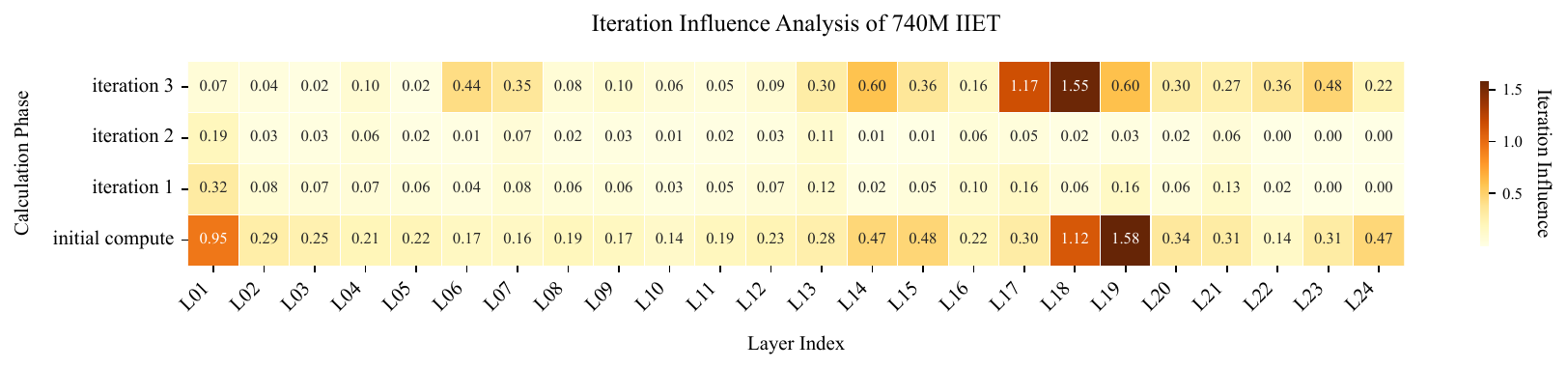}
  \caption{Impact of different iteration stages on the hidden state within each layer of the 740M IIET model, which we term \textbf{iteration influence}. Deeper colors indicate larger hidden state changes after this iteration.}
  \vspace{-3mm}
  \label{fig:fine_grained_bi_heatmap_740m}
\end{figure*}

\begin{table}[t]
    \centering
    \scalebox{0.53}{  
    \begin{tabular}{l cccccccc}
    \toprule
    \textbf{Model} & {\textbf{Para.}} & {\textbf{LMB.}} & {\textbf{PiQA}} & {\textbf{Hella.}} & {\textbf{SCIQ}} & {\textbf{ARC-c}} & {\textbf{Wino.}} & {\textbf{Avg.}} \\
    \midrule
    IIET & 340M & 37.1 & 65.2 & 36.9 & 79.4 & 23.9 & 51.0 & 48.9 \\
    Transformer++ & 1.3B & 37.3 & 65.7 & 37.6 & 78.6 & 23.7 & 51.5 & 49.0 \\
    \midrule
    Quantization (4-bit) & 1.3B & 35.9 & 65.3 & 36.5 & 77.3 & 22.5 & 49.3 & 47.8 \\
    \midrule
    Prun3 & 1.1B & 23.3 & 62.5 & 34.2 & 72.8 & 23.0 & 50.0 & 44.3 \\
    Prun6 & 975M & 11.5 & 59.4 & 32.5 & 67.7 & 22.7 & 50.4 & 40.7 \\
    Prun12 & 650M & 2.9 & 53.3 & 28.9 & 52.5 & 22.1 & 50.9 & 35.1 \\
    \bottomrule
    \end{tabular}}
    \caption{Performance comparison between IIET and compressed variants of the vanilla Transformer. The Transformer variants were created using pruning and quantization to match the storage footprint of IIET.}
    \vspace{-3mm}
    \label{tab:comparison_compress}
\end{table}

\paragraph{Pruning} We use the method from ShortGPT~\cite{men2024shortgpt} to perform layer-wise pruning experiments on the vanilla Transformer. We calculate the block influence for each layer and create three pruned models (Prun3, Prun6, and Prun12) by removing the 3, 6, and 12 layers with the lowest influence scores, respectively. However, we find that pruning has a significant impact on the performance of Transformer++. As shown in Table~\ref{tab:comparison_compress}, compared to the original model, the accuracy of Prun3 decreases by 4.6 points, while Prun6 shows a more substantial drop of 8.2 points. When half of the layers are pruned (Prun12), the model's performance on all test sets was nearly random, so we did not further increase the pruning ratio. Notably, all pruned models performed worse than the 340M IIET model.

\section{Iteration Influence of 740M IIET}
\label{sec:740m_iiet_heatmap}
Figure~\ref{fig:fine_grained_bi_heatmap_740m} displays the Iteration Influence of the 740M IIET model. By selecting the minimum initial computation of each layer as the threshold, we can reduce the number of corrective iterations from 72 to 23.

\section{Two-Stage Training for IIET}

Motivated by the significant computational expense of pre-training large language models from scratch, we investigated a two-stage training paradigm for transitioning a vanilla Transformer to an IIET. Our methodology draws inspiration from the learning rate strategies of recent large models like MiniCPM~\cite{hu2024minicpm} and Deepseek-v3~\cite{liu2024deepseek}, which characteristically maintain a constant learning rate after an initial warmup period during pre-training. A key advantage of this stable training phase is that it facilitates dynamic adjustments to the data curriculum. In this work, we extend this principle of in-training adaptation from the data to the model itself by transitioning the model architecture from a vanilla Transformer to an IIET during the stable learning rate phase.

Our training methodology is partitioned into two distinct phases: \ding{182} \textbf{Foundational Pretraining ($\approx$ 75\% of tokens).} We first pre-train a standard Transformer++ using a constant learning rate of 3e-4, preceded by a 300-step linear warmup. This stage efficiently establishes a robust feature foundation while avoiding the IIET's computational overhead. \ding{183} \textbf{Architectural Transition ($\approx$ 25\% of tokens).} The model's architecture is then transitioned to an IIET for the remaining training. The learning rate decays from 3e-4 via a cosine schedule, leveraging the IIET's dynamics for final model refinement. The experiments are conducted at the 740M parameter scale, see Appendix~\ref{sec:two_stage_detail} for detailed experimental settings.

As presented in Table~\ref{tab:two_stage_results}, our two-stage IIET (denoted ``Two-stage'') demonstrates impressive performance. A significant performance gain is observed during the second training phase, ultimately achieving results comparable to an IIET trained from scratch (51.5 vs. 51.9). Furthermore, this approach significantly outperforms the vanilla Transformer (Trans++) trained on the same data. By introducing IIET only during the final quarter of the training process, the associated computational overhead is reduced by 75\%. This finding offers a promising path for applying IIET to models at much larger scales, and we anticipate that future work will continue to explore this direction.

\begin{table}[t]
    \centering
    \scalebox{0.62}{  
    \begin{tabular}{l ccccccc} 
    \toprule
    \textbf{Model} & {\textbf{LMB.}} & {\textbf{PiQA}} & {\textbf{Hella.}} & {\textbf{SCIQ}} & {\textbf{ARC-c}} & {\textbf{Wino.}} & {\textbf{Avg.}} \\
    \midrule
    Transformer++ & 36.1 & 66.4 & 38.4 & 78.6 & 24.5 & 50.2 & 49.0 \\
    PCformer & 41.0 & 66.3 & 41.3 & 82.0 & 23.3 & 51.2 & 50.9 \\
    IIET & 41.2 & 68.9 & 42.5 & 82.1 & 23.8 & 53.1 & 51.9\\
    \midrule
    Two-stage & 39.3 & 67.5 & 41.3 & 81.9 & 25.3 & 53.7 & 51.5 \\
    \bottomrule
    \end{tabular}}
    \caption{Performance comparison of the two-stage IIET and baseline models at the 740M parameter scale.}
    \vspace{-3mm}
    \label{tab:two_stage_results}
\end{table}

\end{document}